\documentclass{article}

\PassOptionsToPackage{numbers, compress}{natbib}

\usepackage[preprint]{neurips_2020}

\usepackage[utf8]{inputenc} 
\usepackage[T1]{fontenc}    
\usepackage{url}            
\usepackage{booktabs}       
\usepackage{amsfonts}       
\usepackage{nicefrac}       
\usepackage{microtype}      
\usepackage{mathtools}
\usepackage{subcaption}

\DeclarePairedDelimiterX{\infdivx}[2]{(}{)}{%
	#1\;\delimsize\|\;#2%
}

\def\BState{\State\hskip-\ALG@thistlm}
\makeatother

\title{MarsExplorer: Exploration of Unknown Terrains via Deep Reinforcement Learning and Procedurally Generated Environments}

\author{
	Dimitrios I. Koutras$^{1,2}$\\
	\And
	Athanasios Ch. Kapoutsis$^{2, }$\thanks{Corresponding author} \\
	\And
	Angelos A. Amanatiadis$^{3}$\\
	\And
	Elias B. Kosmatopoulos$^{1,2}$\\
	$^{1}$Department of Electrical and Computer Engineering, Democritus University of Thrace\\
	$^{2}$Information Technologies Institute, The Centre for Research \& Technology, Hellas\\
	 $^{3}$Department of Production and Management Engineering\\
}

\begin{document}
	
	\maketitle
	
	\begin{abstract}
	This paper is an initial endeavor to bridge the gap between powerful Deep Reinforcement Learning methodologies and the problem of exploration/coverage of unknown terrains. Within this scope, MarsExplorer, an openai-gym compatible environment tailored to exploration/coverage of unknown areas, is presented. MarsExplorer translates the original robotics problem into a Reinforcement Learning setup that various off-the-shelf algorithms can tackle. Any learned policy can be straightforwardly applied to a robotic platform without an elaborate simulation model of the robot's dynamics to apply a different learning/adaptation phase. One of its core features is the controllable multi-dimensional procedural generation of terrains, which is the key for producing policies with strong generalization capabilities. Four different state-of-the-art RL algorithms (A3C, PPO, Rainbow, and SAC) are trained on the MarsExplorer environment, and a proper evaluation of their results compared to the average human-level performance is reported. In the follow-up experimental analysis, the effect of the multi-dimensional difficulty setting on the learning capabilities of the best-performing algorithm (PPO) is analyzed. A milestone result is the generation of an exploration policy that follows the Hilbert curve without providing this information to the environment or rewarding directly or indirectly Hilbert-curve-like trajectories. The experimental analysis is concluded by evaluating PPO learned policy algorithm side-by-side with frontier-based exploration strategies. A study on the performance curves revealed that PPO-based policy was capable of performing adaptive-to-the-unknown-terrain sweeping without leaving expensive-to-revisit areas uncovered, underlying the capability of RL-based methodologies to tackle exploration tasks efficiently. The source code can be found at: https://github.com/dimikout3/MarsExplorer.
	\end{abstract}

		\section{Introduction}
	
	\subsection{Motivation}
	At this very moment, three different uncrewed spaceships, PERSEVERANCE (USA), HOPE (UAE), TIANWEN-1 (China), are in the surface or in the orbit of Mars. Never before such a diverse array of scientific gear had arrived at a foreign planet at the same time, and with such broad ambitions \cite{witze2020all}. On top of that, several lunar missions have been arranged for this year to enable extensive experimentation, investigation, and testing on an extraterrestrial body \cite{smith2020artemis}. In this exponentially growing field of extraterrestrial missions, a task of paramount importance is the autonomous exploration/coverage of previously unknown areas. The effectiveness and efficiency of such autonomous explorers may significantly impact the timely accomplishment of crucial tasks (e.g., before the fuel depletion) and, ultimately, the success (or not) of the overall mission.
	
	Exploration/coverage of unknown territories is translated into the online design of the path for the robot, taking as input the sensory information and having as objective to map the whole area in the minimum possible time \cite{shrestha2019learned} \cite{kapoutsis2016real}. This setup shares the same properties and objectives with the well-known NP-complete setup of Traveling Salesman Problem (TSP), with the even more restrictive property that the area to be covered is discovered incrementally during the operation.
	
	\subsection{Related Work}
	The well-established family of approaches incorporates the concept of \textit{next best pose} process, i.e. a turn-based, greedy selection of the next best position (also known as \textit{frontier-cell}) to acquire measurement, based on heuristic strategy (e.g., \cite{batinovic2021multi}, \cite{renzaglia2019combining}, \cite{basilico2011exploration}). Although this family of approaches has been extensively studied, some inherent drawbacks significantly constrain its broader applicability. For example, every deadlock that may arise during the previously described optimization scheme should have been predicted, and a corresponding mitigation plan should have been already in place \cite{palacios2016distributed}; otherwise, the robot is going to be stuck in this locally optimal configuration \cite{koutras2020autonomous}. On top of that, to engineer a multi-term strategy that reflects the task at hand is not always trivial \cite{popov2017data}.
	
	The recent breakthroughs in Reinforcement Learning (RL), in terms of both algorithms and hardware acceleration, have spawned methodologies capable of achieving above human-level performance in high-dimensional, non-linear setups, such as the game of Go \cite{silver2016mastering}, atari games \cite{mnih2015human}, multi-agent collaboration \cite{baker2019emergent}, robotic manipulation \cite{zhu2019ingredients}, etc. A milestone in the RL community was the standardization of several key problems under a common framework, namely openai-gym \cite{brockman2016openai}. Such release eased the evaluation among different methodologies and ultimately led to the generation of a whole new series of RL frameworks with standardized algorithms (e.g., \cite{baselines}, \cite{liang2018rllib}), all tuned to tackle openai-gym compatible setups.
	
	These breakthroughs motivated the appliance of RL methodologies in the path-planning/ exploration robotic tasks. Initially, the problem of navigating a single robot in previously unknown areas to reach a destination, while simultaneously avoiding catastrophic collisions, was tackled with RL methods \cite{lei2018dynamic}, \cite{wen2020path}, \cite{zhang2018robot}. The first RL methodology solely developed for exploration of unknown areas was developed in \cite{niroui2019deep}, and has successfully presented the potential benefits of RL. Recently, there have been proposed RL methodologies that seek to leverage the deployment of multi-robot systems to cover an operational area \cite{luis2021multiagent}. 
	
	However, \cite{luis2021multiagent} assumes only a single geometry for the environment to be covered, and thus being prone to overfit, rather than being able to generalize in different environments. \cite{niroui2019deep} mitigates this drawback by introducing a learning scheme with 30 different environments during the training phase. Although such a methodology can adequately tackle the generalization problem, the RL agent's performance is still bounded to the diversity of the human-imported environments.

	\subsection{Contributions}
	The main contribution of this work is to provide a framework for learning exploration/coverage policies that possess strong generalization abilities due to the procedurally generated terrain diversity. The intuition behind such an approach to exploration tasks is the fact that most areas exhibit some kind of structure in their terrain topology, e.g., city blocks, trees in a forest, containers in ports, office complexes. Thereby, by training multiple times in such correlated and procedurally generated environments, the robot will grasp/understand the underlining structure and leverage it to efficiently complete its goal, even in areas that it has never been exposed to.           
	
	Within this scope, a novel openai-gym compatible environment for exploration/coverage of unknown terrains has been developed and is presented. All the core elements that govern a real exploration/coverage setup have been included. MarsExplorer is one of the few RL environments where any learned policy can be transferred to real-world robotic platforms, providing that a proper translation between the proprioceptive/exteroceptive sensors' readings and the generation of 2D perception (occupancy map), as depicted in figure \ref{fig:observation_explain}, and also an integration with the existing robotic systems (e.g., PID low level control, safety mechanisms, etc.) are implemented.
	
	Four state-of-the-art RL algorithms, namely A3C \cite{mnih2016asynchronous}, PPO \cite{schulman2017proximal}, Rainbow \cite{hessel2018rainbow} and SAC \cite{haarnoja2018soft}, have been evaluated on MarsExplorer environment. To better comprehend these evaluation results, the average human-level performance in the MarsExplorer environment is also reported. A follow-up analysis utilizing the best-performing algorithm (PPO) is conducted with respect to the different levels of difficulty. The visualization of the produced trajectories revealed that the PPO algorithm had learned to apply the famous space-filling Hilbert curve, with the additional capability of avoiding on-the-fly obstacles that might appear on the terrain. The analysis is concluded with a scalability study and a comparison with non-learning methodologies.
	
	It should be highlighted that the objective is not to provide another highly realistic simulator but a framework upon which RL methods (and also non-learning approaches) will be efficiently benchmarked in exploration/coverage tasks. Although there are available several wrappers for high-fidelity simulators (e.g. Gazebo \cite{zamora2016extending}, ROS \cite{DBLP:journals/corr/abs-1903-06278}) that could be tuned to formulate an exploration coverage setup, in practice the required execution time for each episode severely limits the type of algorithms that can be used (for example PPO usually needs several millions of steps to environment interactions to converge). To the best of our knowledge, this is the first openai-gym compatible framework oriented for robotic exploration/coverage of unknown areas.
	
	\begin{figure}[h]
		\centering
		\begin{subfigure}[b]{0.42\textwidth}
			\centering
			\includegraphics[width=\textwidth]{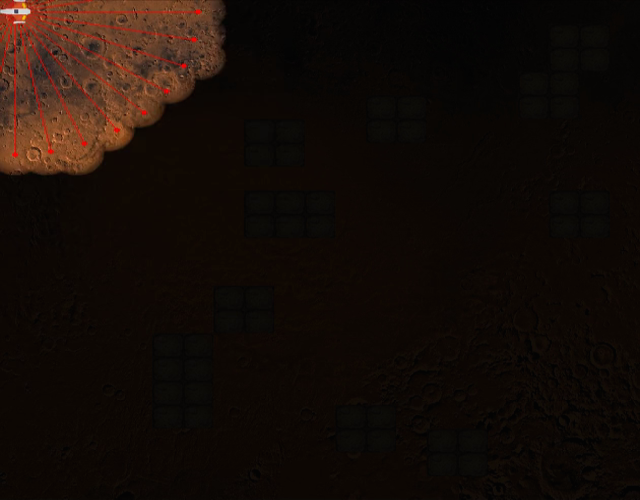}
			\caption{Initial timestep}
			\label{fig:intro_fig_1}
		\end{subfigure}
		\begin{subfigure}[b]{0.42\textwidth}
			\centering
			\includegraphics[width=\textwidth]{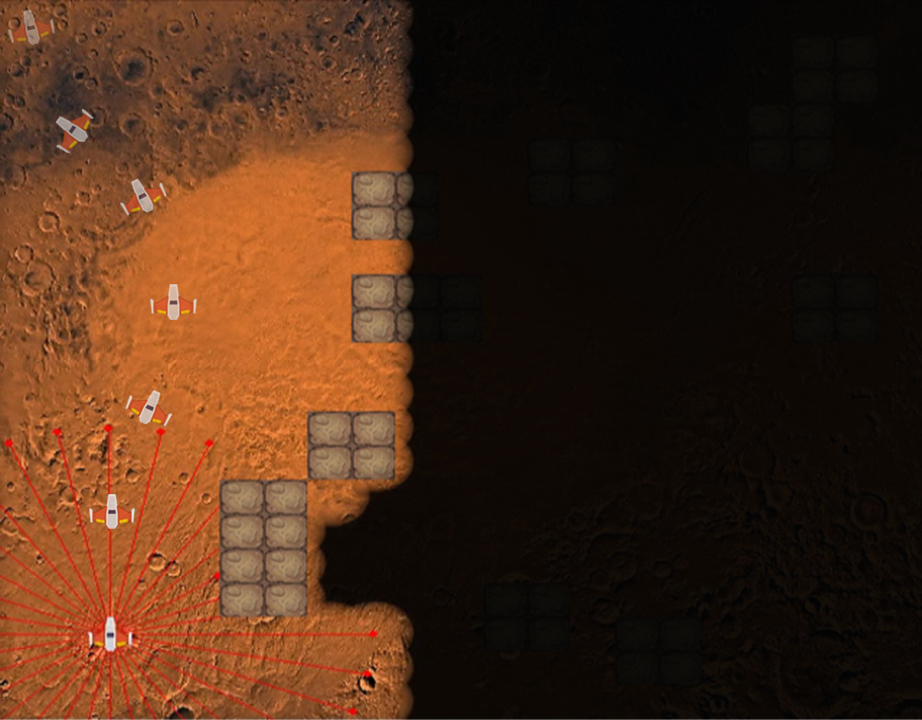}
			\caption{30\% progress}
			\label{fig:intro_fig_2}
		\end{subfigure}
		\begin{subfigure}[b]{0.42\textwidth}
			\centering
			\includegraphics[width=\textwidth]{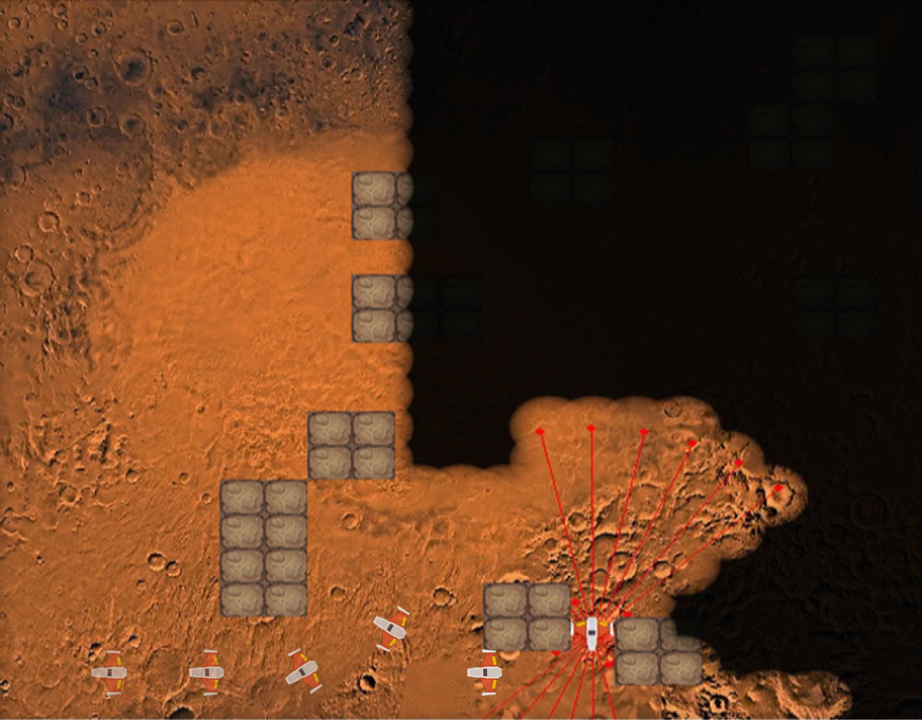}
			\caption{65\% progress}
			\label{fig:intro_fig_3}
		\end{subfigure}
		\begin{subfigure}[b]{0.42\textwidth}
			\centering
			\includegraphics[width=\textwidth]{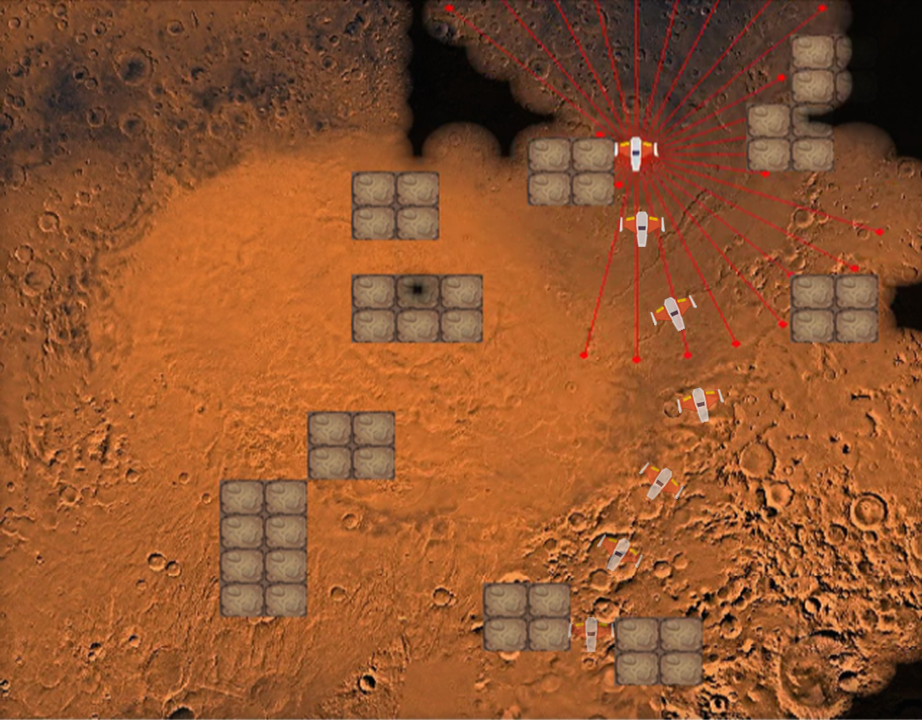}
			\caption{Final timestep}
			\label{fig:intro_fig_4}
		\end{subfigure}
		\caption{Indicative example: Trained RL agent executes exploration/coverage task in previously unknown and cluttered terrain utilizing MarsExplorer environment.}
		\label{fig:intro_fig}
	\end{figure}
	
	Figure \ref{fig:intro_fig} presents 4 sample snapshots that illustrate the performance of a trained RL robot inside the MarsExplorer environment. Figure \ref{fig:intro_fig_1} demonstrates the robot's entry inside the unknown terrain, which is annotated with black color. Figure \ref{fig:intro_fig_2} illustrates all the so-far gained ``knowledge'', which is either depicted with Martian soil or brown boxes to denote free space or obstructed positions, respectively. An attractive trait is depicted in figure \ref{fig:intro_fig_3}, where the robot chose to perform a dexterous maneuver between two obstacles to be as efficient as possible in terms of numbers of timesteps for the coverage task. Note that any collision with an obstacle would have resulted in a termination of the episode and, as a result, an acquisition of an extreme negative reward. Figure \ref{fig:intro_fig_4} illustrates the robot's final position, along with all the gained information for the terrain (non-black region) during the episode.
	
	All in all, the main contributions of this paper are:
	\begin{itemize}
		\item Develop an open-source\footnote{https://github.com/dimikout3/MarsExplorer}, openai-gym compatible environment tailored explicitly to the problem of exploration of unknown areas with an emphasis on generalization abilities.
		\item Translate the original robotics exploration problem to an RL setup, paving the way to apply off-the-shelf algorithms.
		\item Perform preliminary study on various state-of-the-art RL algorithms, including A3C, PPO, Rainbow, and SAC, utilizing the human-level performance as a baseline.
		\item Challenge the generalization abilities of the best performing PPO-based agent by evaluating multi-dimensional difficulty settings.
		\item Present side-by-side comparison with frontier-based exploration strategies.
	\end{itemize}
	
	\subsection{Paper outline}
	The remaining of this paper is organized as follows: Section \ref{sec:environment} presents the details of the openai-gym exploration environment, called MarsExplorer, along with an analysis of the key RL attributes inserted. Section \ref{sec:results} presents the experimental analysis from the survey regarding the performance of the state-of-the-art RL algorithm to the evaluation against standard frontier-based exploration. Finally, section \ref{sec:conclusions} summarizes the findings the draws the conclusions of this study.
	
	\section{Environment}
	\label{sec:environment}
	
	This section identifies the fundamental elements that govern the family of setups that fall into the coverage/exploration class and translates them to the openai gym framework \cite{brockman2016openai}. In principle,  the objective of the robot is to cover an area of interest in the minimum possible time while avoiding any non-traversable objects, the position of which gets revealed only when the robot's position is in close proximity \cite{kapoutsis2019distributed}, \cite{burgard2000collaborative}.
	
	\subsection{Setup}
	\label{subsec:Setup}
	Let us assume that area to be covered is constrained within a square, which has been discretized into $n = rows \times cols$ identical grid cells:
	\begin{equation}
		\label{eq:environment}
		{\cal G} = \left\lbrace (x,y): x \in [1, rows], y \in [1, cols] \right\rbrace
	\end{equation}
	
	The robot cannot move freely inside this grid, as some grid cells are occupied by non-traversable objects (obstacles). Therefore, the map of the terrain is defined as follows:
	\begin{equation}
		\label{eq:map}
		{\cal M}\left(q \right)  =  \left\{ \begin{array}{ll}
			0.3 & \mbox{\textit{free space}} \\
			1 & \mbox{\textit{obstacle}}
		\end{array}\right. \; q=(x, y) \in {\cal G}
	\end{equation}
	
	The values of ${\cal M}$ correspond to the morphology of the unknown terrain and are considered a priori unknown.  
	
	\subsection{Action space}
	\label{subsec:action_space}
	Keeping in mind that the movement capabilities of the robot mainly impose the discretization of the area into grid cells, the action space is defined in the same grid context as well. The position of the robot is denoted by the corresponding $x, y$ cell of the grid, i.e. $p_a(t) = \left[x_a(t), y_a(t)\right] $. Then, the possible next actions are simply given by the Von Neumann neighborhood \cite{gray2003mathematician}, i.e.
	\begin{equation}
		\label{eq:action_space}
		{\cal A}_{p_a} = \left\lbrace (x,y): \left|x-x_a\right| +\left|y-y_a\right| \leq 1 \right\rbrace
	\end{equation}
	
	In the openai-gym framework, the formulation above is realized by a discrete space of size 4 (North, East, South, West).
	
	\subsection{State space}
	\label{subsec:state_space}
	With each movement, the robot may acquire some information related to the formation of the environment that lies inside its sensing capabilities, according to the following lidar-like model:
	\begin{equation}
		\label{eq:lidarModel}
		y_q(t) = \left\{ \begin{array}{ll} 1 & \mbox{ if } \left\|p_a(t) - q \right\| \leq d \mbox{ AND  } \\
			& \exists \mbox{ line-of-sight between} \\
			&  p_a(t) \mbox{ and } q  \\
			0 & \mbox{ otherwise}
		\end{array}\right. \; \forall q \in {\cal G} \
	\end{equation}
	
	where $d$ denotes the maximum scanning distance.
	
	An auxiliary boolean matrix $D(t)$ is introduced to denote all the cells that have been discovered from the beginning till $t$ timestep. $D(t)$ annotates with one all cells that have been sensed and with zero all the others. Starting from a zero matrix $rows \times cols$, its values are updated as follows:
	\begin{equation}
		\label{eq:visibleMatrix}
		D_q(t) = D_q(t-1) \lor y_q(t), \;\;\;\; \forall q \in {\cal G}
	\end{equation}
	
	where $\lor$ denotes the logical OR operator. The state is simply an aggregation of the acquired information over all past measurements of the robot (\ref{eq:lidarModel}). Having updated (\ref{eq:visibleMatrix}), the state $s(t_k)$ is a matrix of the same size as the grid to be explored (\ref{eq:environment}), where its values are given by:
	
	\begin{equation}
		\label{eq:observationModel}
		s_q(t) = \left\{ \begin{array}{ll} {\cal M}_q & \mbox{ if } D_q(t)
			\\
			0 \; (=\mbox{\textit{undefined}}) &  \mbox{otherwise}
		\end{array}\right. \; \forall q \in {\cal G} \
	\end{equation}
	
\begin{figure}[!h]
		\centering
		\begin{subfigure}[b]{0.42\textwidth}
			\centering
			\includegraphics[width=\textwidth]{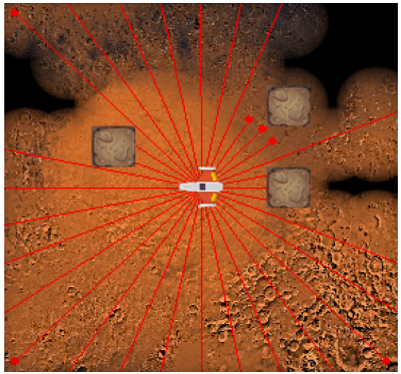}
			\caption{Graphical environment}
			\label{fig:observation_explain_1}
		\end{subfigure}
		\hfill
		\begin{subfigure}[b]{0.42\textwidth}
			\centering
			\includegraphics[width=\textwidth]{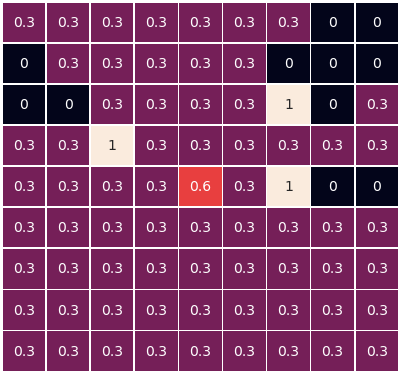}
			\caption{State $s(t)$ representation }
			\label{fig:observation_explain_2}
		\end{subfigure}
		\caption{State encoding}
		\label{fig:observation_explain}
	\end{figure}

	Finally, the robot's position is declared by making the value of the corresponding cell equal to $0.6$, i.e. $s_{q = p_a(t)}(t) = 0.6$. Overall, state $s(t)$ is defined as a 2D matrix, that takes values from the following discrete set: $\{0,0.3,0.6,1\}$. Figure \ref{fig:observation_explain} presents an illustrative example of a registration between the graphical environment (figure \ref{fig:observation_explain_1}) and the corresponding state representation (figure \ref{fig:observation_explain_2}).
	
	\subsection{Reward function}
	\label{subsec:reward}
	Having in mind that the ultimate objective is to discover all grid cells, the instantaneous core reward, at each timestep $t$, is defined as the number of newly explored cells, i.e.
	\begin{equation}
		\label{eq:core_reward}
		r_{explor}(t) = \sum_{q \in {\cal G}} D_q(t) - \sum_{q \in {\cal G}} D_q(t-1)
	\end{equation}
	Intuitively, if $\sum_{k=0}^{T} r_{explor}(k) \to n$ , then the robot has explored the whole grid (\ref{eq:environment}) in $T$ timesteps.
	
	To force robot to explore the whole area (\ref{eq:core_reward}), while avoiding unnecessary movements, an additional penalty $r_{move}=0.5$ per timestep is applied. In essence, this negative reward aims to distinguish among policies that lead to the same number of discovered cells but needed a different number of exploration steps. Please note that the value of $r_{move}$ should be less than $1$, to have less priority than the exploration of a single cell.
	
	The action space, as defined previously, may include invalid next movements for the robot, i.e., out of the operational area (\ref{eq:environment}) or crashing into a discovered obstacle. Thus, apart from the problem at hand, the robot should be able to recognize these undesirable states and avoid them at all costs. Towards that direction, an additional penalty $r_{invalid} = n$ is introduced for the cases where the next robot's movement leads to an invalid state. Along with such a reward, the episode is marked as ``done'', indicating that a new episode should be initiated.
	
	At the other side of the spectrum, a completion bonus $r_{bonus} = n$ is given to the robot when more than $\beta$\% (e.g., 95\%) of the cells have been explored. Similar to the previous case, this is also considered a terminal state.
	
	Putting everything together, the reward is defined as:
	\begin{equation}
		\label{eq:reward}
		r(t) =  \left\{ \begin{array}{ll} -r_{invalid} & \mbox{if next state} \\
			&  \mbox{is invalid}\\
			r_{explor}(t) - r_{move} +  &\\
			\left\{ \begin{array}{ll} r_{bonus} & \mbox{if } \frac{\sum_{q \in {\cal G}} D(t)}{n} \geq \beta \\
				0 & \mbox{otherwise} \end{array}\right. & \mbox{otherwise}
		\end{array}\right. \;
	\end{equation}
	
	\subsection{Key RL Attributes} 
	\label{subsec:rl_problem}
	MarsExplorer was designed as an initial endeavor to bridge the gap between powerful existing RL algorithms and the problem of autonomous exploration/coverage of a previously unknown, cluttered terrain. This subsection presents the build-in key attributes of the designed framework.
	
	\textbf{Straightforward applicability.} One of the fundamental attributes of MarsExplorer is that any learned policy can be straightforwardly applied to an appropriate robotic platform with little effort required. This can be achieved by the fact that the policy calculates a high-level exploration path based on the perception of the environment (\ref{eq:observationModel}). Thus, assuming that a smooth integration with the sensor's readings (for example, using a Kalman filter), can be used to represent the environment as in (\ref{eq:observationModel}), no elaborate simulation model of the robot's dynamics is required to adjust the RL algorithm into the specifics of the robotic platform.
	
	\textbf{Terrain Diversity.}
	For each episode, the general dynamics are determined by a specific automated process that has different levels of variation. These levels correspond to the randomness in the number, size, and positioning of obstacles, the terrain scalability (size), the percentage of the terrain that the robot must explore to consider the problem solved, and the bonus reward it will receive in that case. This procedural generation \cite{cobbe2020leveraging} of terrains allows training in multiple/diverse layouts, forcing, ultimately, the RL algorithm to enable generalization capabilities, which are of paramount importance in real-life applications where unforeseen cases may appear.
	
	
	
	\textbf{Partial Observability.}
	Due to the nature of the exploration/coverage setup, at each timestep, the robot is only aware of the location of the obstacles that have been sensed from the beginning of the episode (\ref{eq:visibleMatrix}). Therefore, any long-term plan should be agile enough to be adjusted on the fly, based on future information about the unknown obstacles' positions. Such a property renders the acquisition of a global exploration strategy quite tricky \cite{yin2021sequential}.
	
	\textbf{Fast Evaluation.}
	Disregarding the environment from any irrelevant physics dynamics and focusing only on the exploration/coverage aspect (\ref{eq:environment})-(\ref{eq:reward}), MarsExplorer allows rapid execution of timesteps. This feature can be of paramount importance in the RL ecosystem, where the algorithms usually need millions of timesteps to converge, as it can enable fast experimental pipelines and prototyping.
	
	
	\section{Performance Evaluation}
	\label{sec:results}
	This section presents an experimental evaluation of the MarsExplorer environment. The analysis begins with all the implementation details that are important for realizing the MarsExplorer experimental setup. For the first evaluation iteration, 4 state-of-art RL algorithms are applied and evaluated in a challenging version of MarsExplorer that requires the development of \textit{strong generalization} capabilities in a highly randomized scenario, where the underlying structure is almost absent. Having identified the best performing algorithm, a follow-up analysis is performed with respect to the difficulty vector values. The learned patterns and exploration policies for different evaluation instances are further investigated and graphically presented. The analysis is concluded with a scale-up study in two larger terrains and a comparison between the trained robot and two well-established frontier-based approaches.

	\subsection{Implementation details}
	\label{subsec:implementation_details}
	
	Aside from the standardization as an openai-gym environment, MarsExplorer provides an API that allows manually controlled experiments, translating commands from keyboard arrows to next movements. Such a feature can assess human-level performance in the exploration/coverage problem and reveal important traits by comparing human and machine-made strategies.
	
	Ray/RLlib framework \cite{liang2017ray} was utilized to perform all the experiments. The fact that RLlib is a well-documented, highly-robust library also eases the build-on developments (e.g., apply a different RL pipeline), as it follows a common framework. Furthermore, such an experimental setup may also leverage the interoperability with other powerful frameworks from the Ray ecosystem, e.g., Ray/Tune for hyperparameters' tuning.
	
	\begin{figure}[!h]
		\centering
		\includegraphics[width=0.6\textwidth]{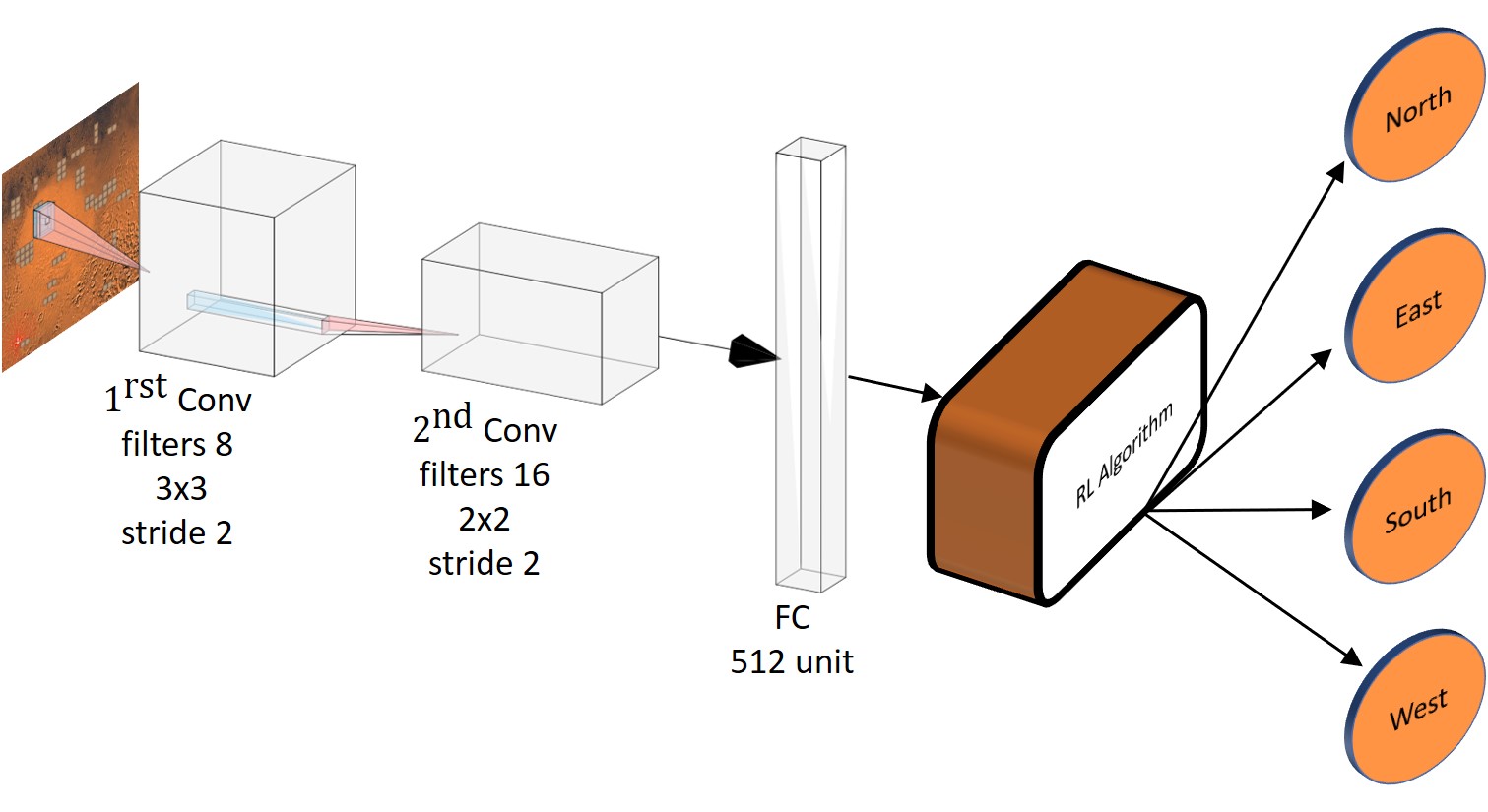}
		\caption{Overview of the experimental architecture}
		\label{fig:General_Architecture}
	\end{figure}
	
	\begin{table}[!h]
		\caption{Implementation parameters}
		\label{tab:parameters}
		\centering
		\begin{tabular}{ c | c  c}
			\hline
			\textbf{Parameter} & Value & Equation\\
			\hline
			Grid size & $[21\times21]$ & (\ref{eq:environment}) \\
			Sensor radius & $d=6$ grid cells  & (\ref{eq:lidarModel}) \\
			Considered done & $\beta = 99\%$ &  (\ref{eq:reward})\\
			\hline
		\end{tabular}
	\end{table}

	Table \ref{tab:parameters} summarizes all the fixed parameters used for all the performed experiments. MarsExplorer admits the distinguishing property of stochastically deploying the obstacles at the beginning of each episode. This stochasticity can be controlled and ultimately determines the difficulty level of the MarsExplorer setup. The state-space of MarsExplorer has a strong resemblance to thoroughly studied 2D environments, e.g., ALE \cite{Bellemare_2013}, only with the key difference that the image is generated incrementally and based on the robot's actions. Therefore, as it has been standardized from the DQN algorithm's application domain \cite{mnih2015humanlevel}, a vision-inspired neural network architecture is incorporated as a first stage. Figure \ref{fig:General_Architecture} illustrates the architecture of this \textit{pre-processor}, which is comprised of 2 convolutional layers followed by a fully connected one. The vectorized output of the fully connected layer is forwarded to a ``controller'' architecture dependent on the RL algorithm enabled.

	\subsection{State-of-the-art RL algorithms comparison}
	\label{subsec:state_of_the_art}
	
	Apart from the details described in the previous subsection, for the comparison study, at the beginning of each episode, the formation (position and shape) of obstacles was set randomly. This choice was made to force RL algorithms to develop novel generalization strategies to tackle such a challenging setup. The list of studied RL algorithms is comprised by the following model-free approaches: PPO \cite{schulman2017proximal}, DQN-Rainbow \cite{hessel2018rainbow}, A3C \cite{mnih2016asynchronous} and SAC \cite{haarnoja2018soft}. All hyperparameters of these algorithms are reported in the Appendix \ref{appendx:Hyperparameters}.
	
	Figure \ref{fig:Comparing_Algorithms} presents a comparison study among the approaches mentioned above. For each RL agent, the thick colored lines stand for the episode's total reward, while the transparent surfaces around them correspond to the standard deviation. Moreover, the episode's reward (\textit{score}) is normalized in such a way that 0 stands for an initial invalid action by the robot, $r_{invalid}$ in (\ref{eq:reward}), while 1 correspond to the theoretical maximum reward, which is the $r_{bonus}$ in (\ref{eq:reward}) plus the number of cells.
	
		\begin{figure}[!h]
			\centering
			\includegraphics[width=0.46\textwidth]{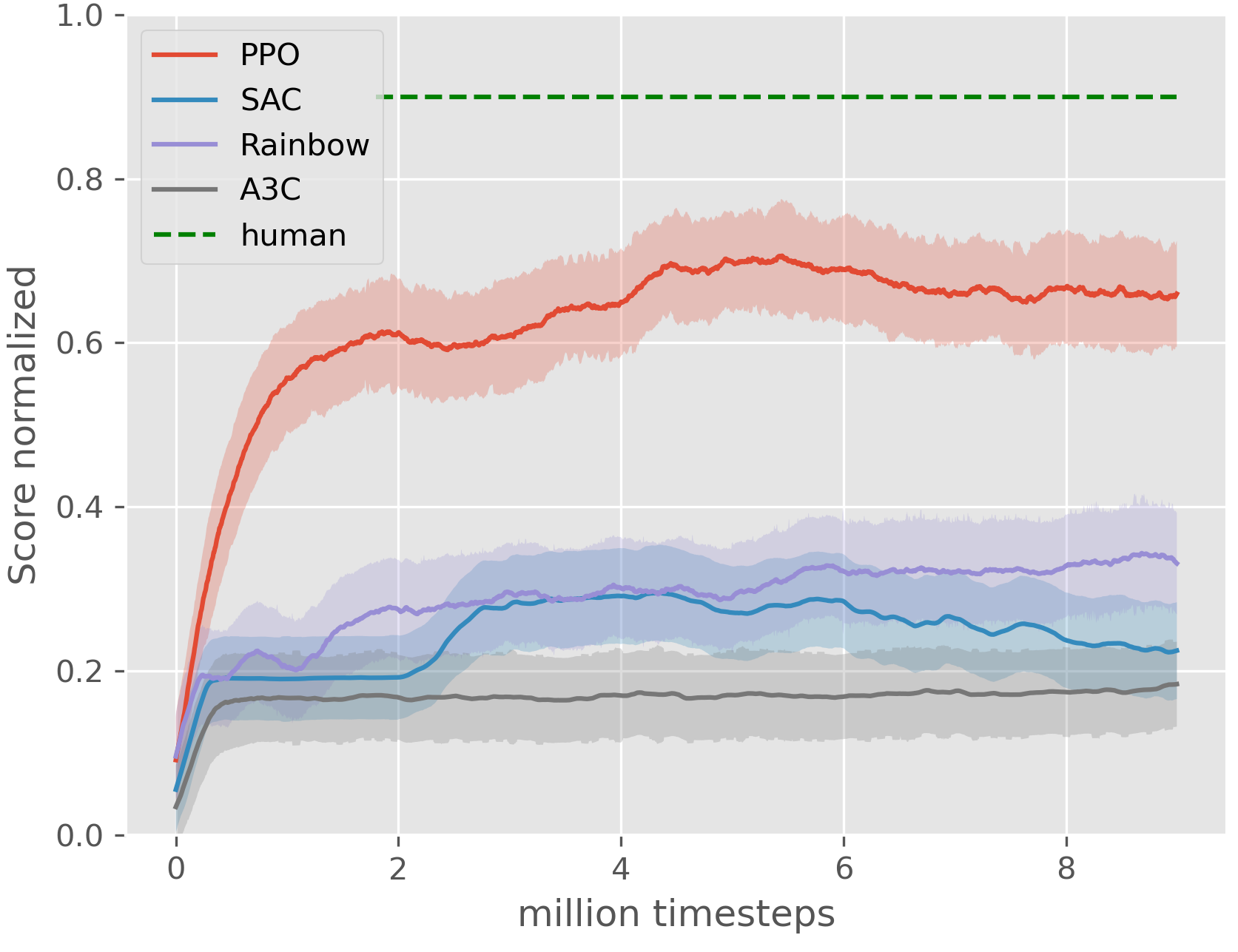}
			\caption{Learning curves for MarsExplorer with randomly chosen obstacles.}
			\label{fig:Comparing_Algorithms}
		\end{figure}
		
		\begin{figure*}[!h]
			\centering
			\begin{subfigure}[b]{0.325\textwidth}
				\centering
				\includegraphics[width=\textwidth]{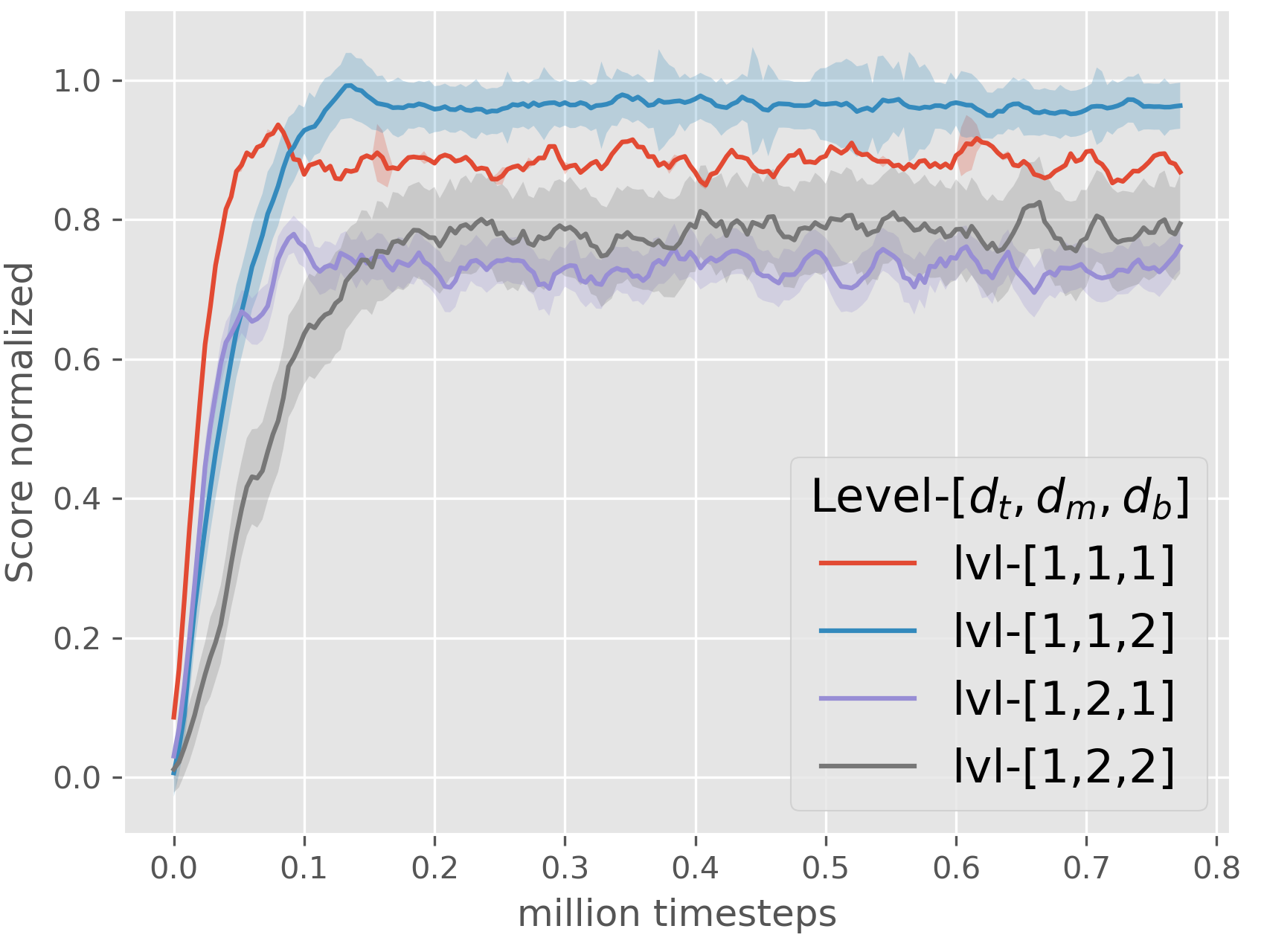}
				\caption{Topology stochasticity level $d_t=1$}
				\label{subfig:topology_stochasticity_1}
			\end{subfigure}
			\hfill
			\begin{subfigure}[b]{0.325\textwidth}
				\centering
				\includegraphics[width=\textwidth]{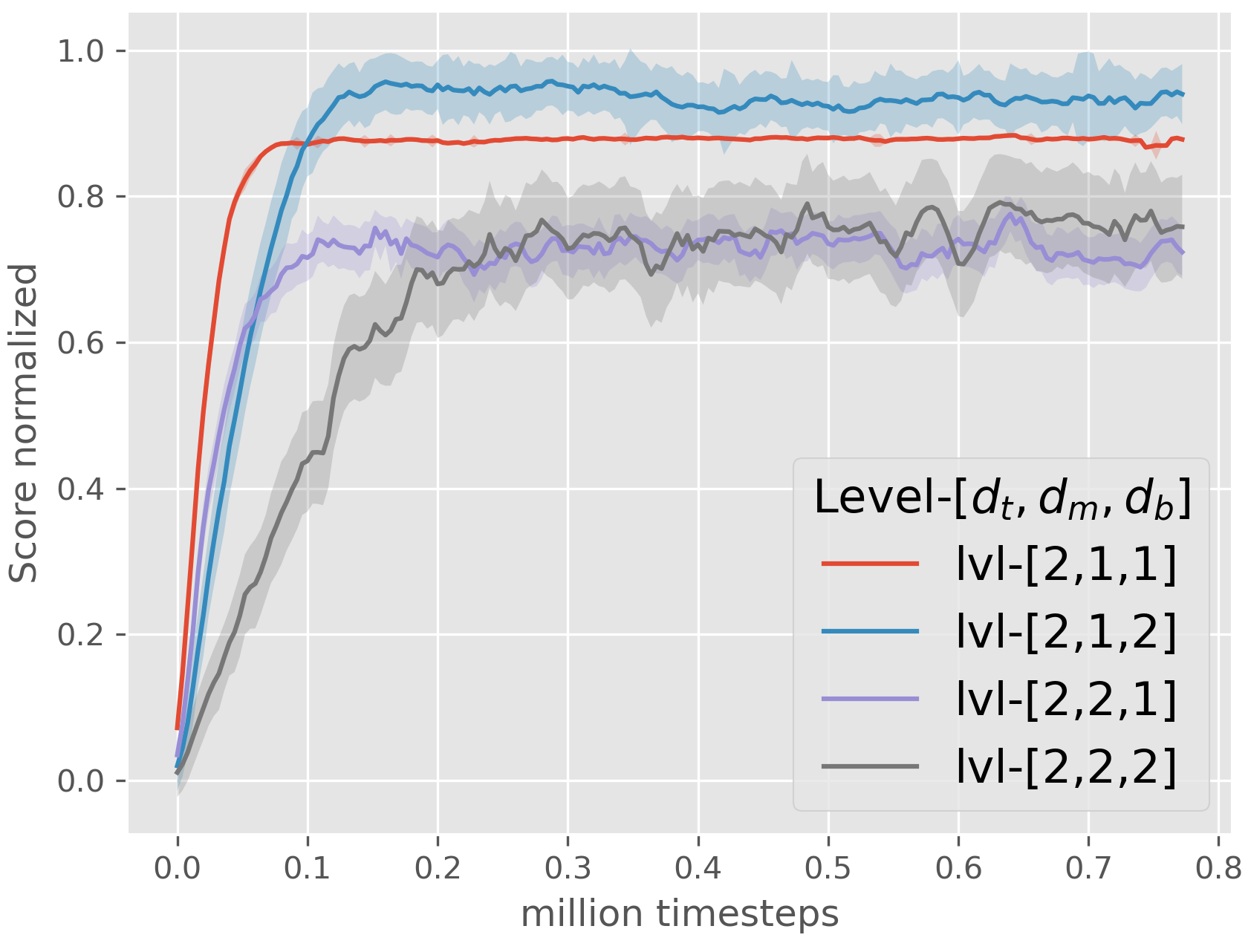}
				\caption{Topology stochasticity level $d_t=2$}
				\label{subfig:topology_stochasticity_2}
			\end{subfigure}
			\begin{subfigure}[b]{0.325\textwidth}
				\centering
				\includegraphics[width=\textwidth]{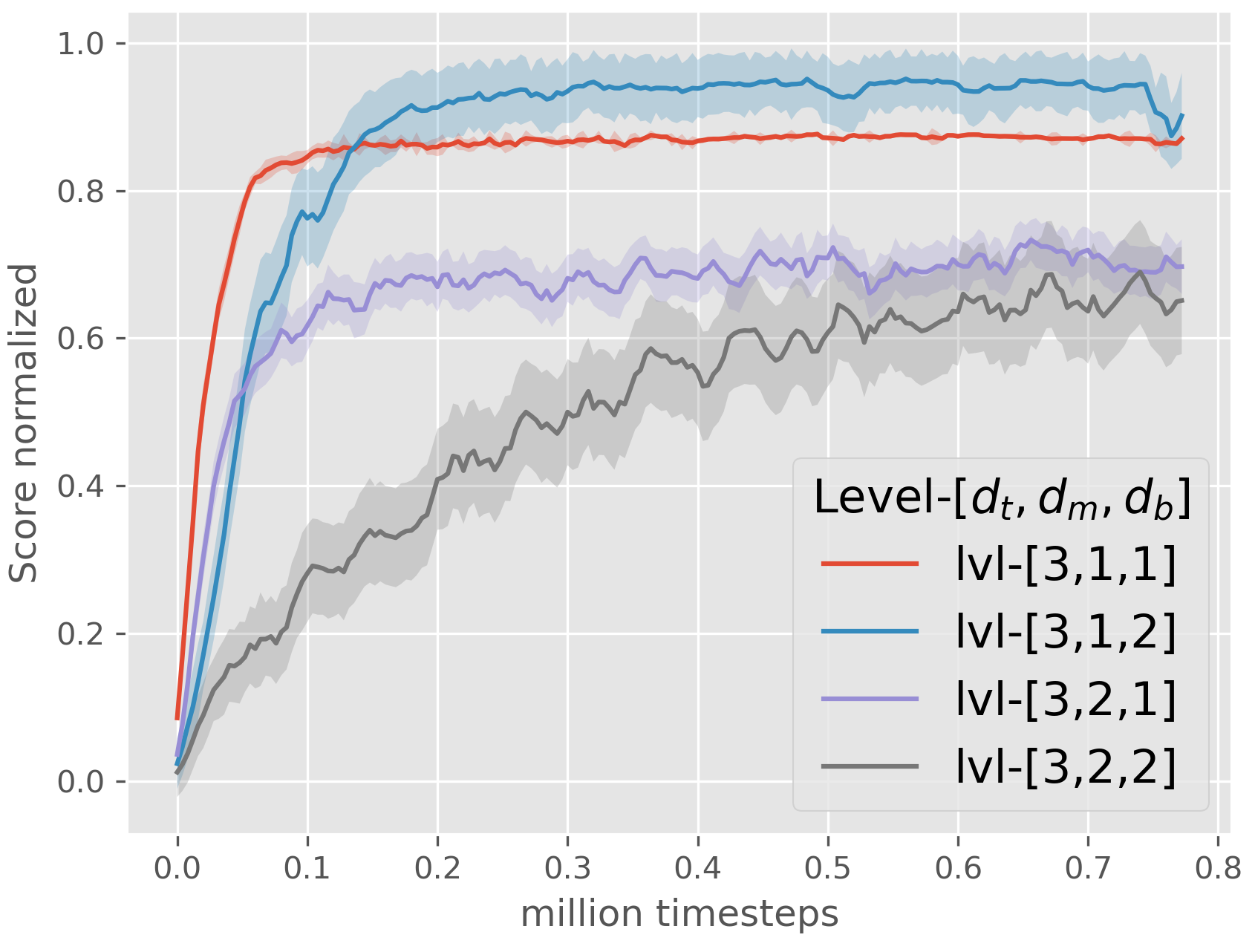}
				\caption{Topology stochasticity level $d_t=3$}
				\label{subfig:topology_stochasticity_3}
			\end{subfigure}
			\caption{The sensitivity of PPO algorithm learning curves with respect to the different levels of multi-dimensional difficulty vector.}
			\label{fig:Dim_Difficulity}
		\end{figure*}
	
	To increase the qualitative comprehension of the produced results, the average human-level performance is also introduced. To approximate this value, 10 players were drawn from the pool of $\textit{CERTH/ConvCAO}$ employees to participate in the evaluation process. Each player had an initial warm-up phase of 15 episodes (non-ranked), and after that, they were evaluated on 30 episodes. The average achieved score of the 300 human-controlled experiments is depicted with a green dashed line.
	
	A clear-cut outcome is that the PPO algorithm achieves the highest average episodic reward, reaching an impressive 85.8\% of the human-level performance. DQN-Rainbow achieves the second-best performance; however, the average is 50.04\% and 42.73\% of the PPO and human-level performance, respectively.
	
	\subsection{Multi-dimensional difficulty}
	\label{subsec:multi_dimensional_difficulty}
	
	Having defined the best performing RL algorithm (PPO), now the focus is shifted on producing some preliminary results, related with the difficulty settings of MarsExplorer. As mentioned in the definition section, MarsExplorer allows for setting the elements of difficulty vector independently. More specifically, the difficulty vector comprised of 3 elements $[d_t, d_m, d_b]$, where:
	
	\textbullet $\;d_t$ denotes the \textbf{topology stochasticity}, which defines the obstacles' placement  on the field. The \textit{fundamental positions} of the obstacles are equally arranged in a 3 columns -- 3 rows format. $d_t$ controls the radius of deviation around these \textit{fundamental positions}. As the value of $d_t$ increases, the obstacles' topology has more unstructured formation. $d_t$ takes values from $\{1, 2, 3\}$ discrete set.
	
	\begin{figure*}[!h]
		\centering
		\begin{subfigure}[b]{0.23\textwidth}
			\centering
			\includegraphics[width=\textwidth]{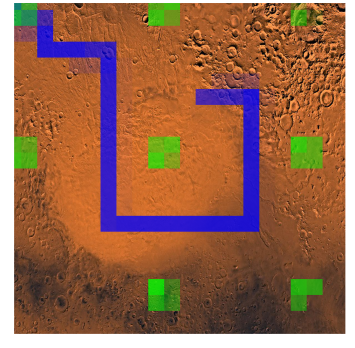}
			\caption{level-[1,1,1]}
			\label{subfig:noise_level_1}
		\end{subfigure}
		\begin{subfigure}[b]{0.23\textwidth}
			\centering
			\includegraphics[width=\textwidth]{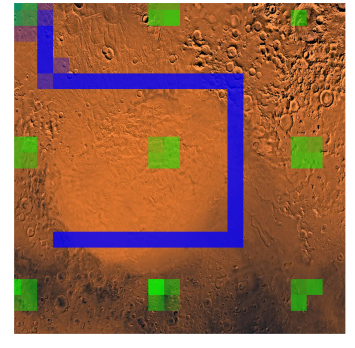}
			\caption{level-[1,1,2]}
			\label{subfig:noise_level_2}
		\end{subfigure}
		\begin{subfigure}[b]{0.23\textwidth}
			\centering
			\includegraphics[width=\textwidth]{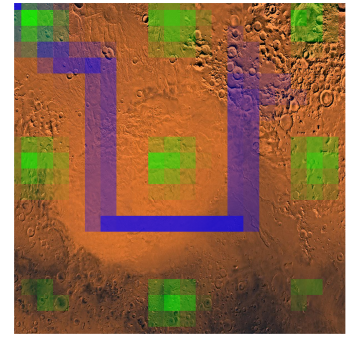}
			\caption{level-[1,2,1]}
			\label{subfig:noise_level_3}
		\end{subfigure}
		\begin{subfigure}[b]{0.23\textwidth}
			\centering
			\includegraphics[width=\textwidth]{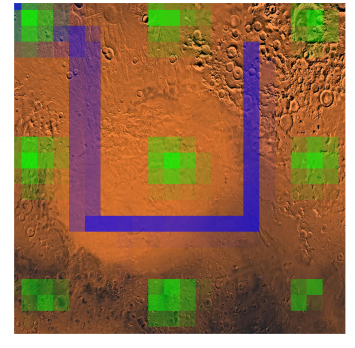}
			\caption{level-[1,2,2]}
			\label{subfig:noise_level_4}
		\end{subfigure}
		\begin{subfigure}[b]{0.23\textwidth}
			\centering
			\includegraphics[width=\textwidth]{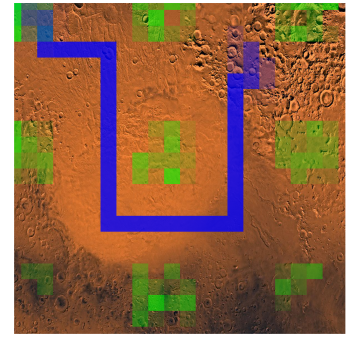}
			\caption{level-[2,1,1]}
			\label{subfig:noise_level_5}
		\end{subfigure}
		\begin{subfigure}[b]{0.23\textwidth}
			\centering
			\includegraphics[width=\textwidth]{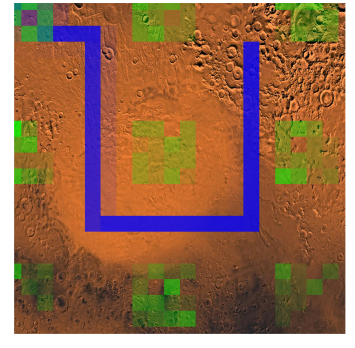}
			\caption{level-[2,1,2]}
			\label{subfig:noise_level_6}
		\end{subfigure}
		\begin{subfigure}[b]{0.23\textwidth}
			\centering
			\includegraphics[width=\textwidth]{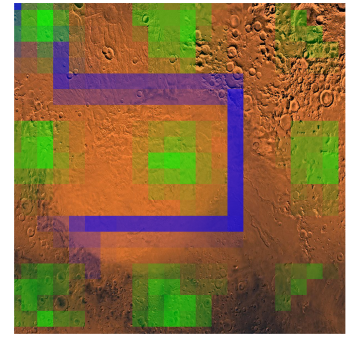}
			\caption{level-[2,2,1]}
			\label{subfig:noise_level_7}
		\end{subfigure}
		\begin{subfigure}[b]{0.23\textwidth}
			\centering
			\includegraphics[width=\textwidth]{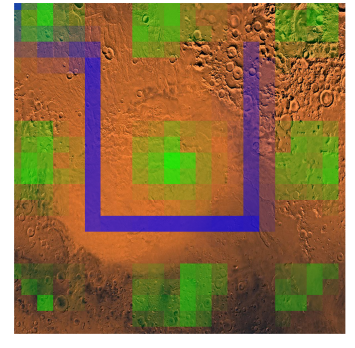}
			\caption{level-[2,2,2]}
			\label{subfig:noise_level_8}
		\end{subfigure}
		\begin{subfigure}[b]{0.23\textwidth}
			\centering
			\includegraphics[width=\textwidth]{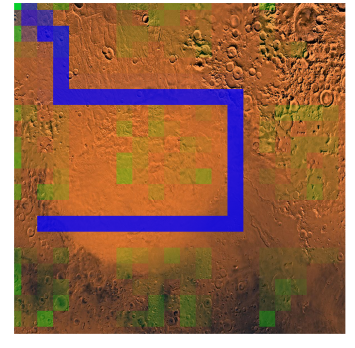}
			\caption{level-[3,1,1]}
			\label{subfig:noise_level_9}
		\end{subfigure}
		\begin{subfigure}[b]{0.23\textwidth}
			\centering
			\includegraphics[width=\textwidth]{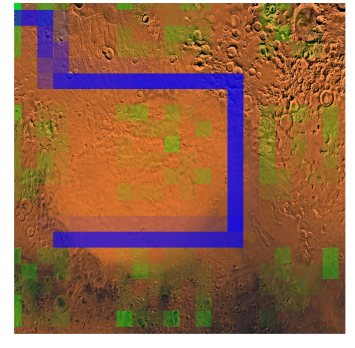}
			\caption{level-[3,1,2]}
			\label{subfig:noise_level_10}
		\end{subfigure}
		\begin{subfigure}[b]{0.23\textwidth}
			\centering
			\includegraphics[width=\textwidth]{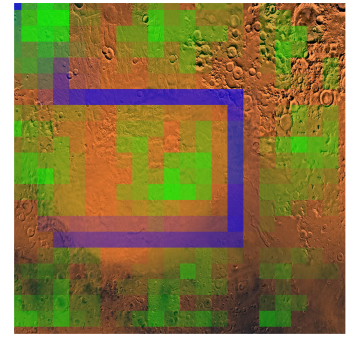}
			\caption{level-[3,2,1]}
			\label{subfig:noise_level_11}
		\end{subfigure}
		\begin{subfigure}[b]{0.23\textwidth}
			\centering
			\includegraphics[width=\textwidth]{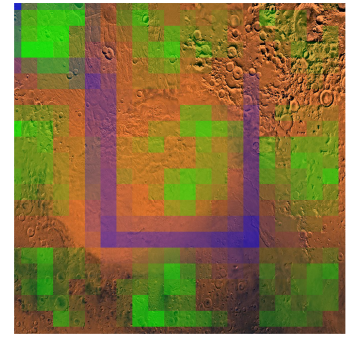}
			\caption{level-[3,2,2]}
			\label{subfig:noise_level_12}
		\end{subfigure}
		\caption{Heatmap of the evaluation results of the learned PPO policy. For each of the 12 difficulty levels, 100 experiments were performed, with the randomness in obstacles' formation as imposed by the corresponding level. Blue colormap corresponds to the frequency of cell visitations by the RL agent, while green colormap corresponds to the location of the encountered obstacles for all the evaluations.}
		\label{fig:heat_maps}
	\end{figure*}
	
	\textbullet $\;d_m$ denotes the \textbf{morphology stochasticity}, which defines the obstacles' shape on the field. $d_m$ controls the area that might be occupied from each obstacle. The bigger the value of $d_m$, the larger the compound areas of obstacles that might appear on the MarsExplorer terrain. $d_m$ takes values from $\{1, 2\}$ discrete set.
	
	\textbullet $\;d_b$ denotes the \textbf{bonus rewards}, that are assigned for the completion ($r_{bonus}$) and failure ($r_{invalid}$) of the mission (\ref{eq:reward}). For this factor only two values are allowed $\{1, 2\}$, that correspond to cases of providing and not-providing the bonus rewards, respectively.
	
	Higher values in the elements of the difficulty vector correspond to less structured behavior in the obstacles formation. Thus, a trained agent that has been successfully trained in greater difficulty setups may exhibit increased generalization abilities. Overall, the aggregation of the aforementioned elements' domain generates 12 combinations of difficulty levels. Figure \ref{fig:Dim_Difficulity} shows the total average return of the evolution of the average episodic reward for each one of the 12 levels during the training of the PPO algorithm. To improve the readability of the graphs, the results are organized into 3 graphs, one for each level of $d_t$, with 4 plot lines each.
	
	A study on the learning curves reveals that $d_m$ has the largest effect on the learned policy. Blue and red lines (cases where $d_m=1$), in all three figures, demonstrate a similar convergence rate and also the highest-performance policies. However, a serious degradation in the results is observed in purple and gray lines ($d_m=2$). As it was expected, when $d_m = 2$ and also $d_t=3$ (purple and gray lines in figure \ref{subfig:topology_stochasticity_3}) the final achieved performance reached only a little bit above 0.6 in the normalized scale. $d_b$ seems that does not affect much the overall performance, at least until this vector of difficulty, apart from the convergence rate depicted in the gray line of figure \ref{subfig:topology_stochasticity_3}.
	
	\subsection{Learned policy evaluation}
	\label{subsec:test}
	
	This section is devoted to the characteristics of the learned policy from the PPO algorithm. For each of the 12 levels of difficulty defined in the previous section, the best PPO policy was extracted and evaluated in a series of 100 experiments with randomly (controlled by the difficulty setting) generated obstacles. Figure \ref{fig:heat_maps} presents one heat map for each difficulty level. Blue colormap corresponds to the frequency of the robot visiting a specific cell of the terrain. Green colormap corresponds to the number of detected obstacles in each position during the robot's exploration.
	
	A critical remark is that, for each scenario, the arrangement of discovered obstacles matches the drawn distribution as described in the previous subsection, implying that the learned policy does not have any ``blind spots''.
	
	Examining the heatmap of the trajectories in each scenario, it is crystal clear that the same family of trajectories has been generated in all cases and with great confidence. The important conclusion here is that this pattern is the first order of the Hilbert curve that has been utilized extensively in the space-filling domain (e.g., \cite{kapoutsis2017darp}, \cite{sadat2015fractal}). Please highlight that such a pattern has not been imported to the simulator or rewarded when achieved from the RL algorithm; however, the algorithm learned that this is the most effective strategy by interacting with the environment.
	
	It would be an omission not to mention the learned policy's ability to adapt to changes in the obstacles' distribution and, ultimately, find the most efficient obstacle-free route. This trait can be observed more clearly in subfigures \ref{subfig:noise_level_11} and \ref{subfig:noise_level_12}, where the policy needed to be extremely dexterous and delicate to avoid obstacles' encounters.

	\subsection{Comparison with frontier-based methodologies for varying terrain sizes}
	\label{subsec:scale}
	
	\begin{figure}[!h]
		\centering
		\includegraphics[width=0.75\textwidth]{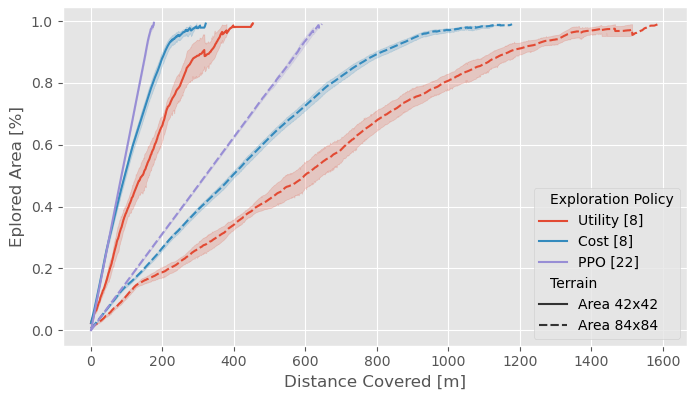}
		\caption{Comparison between 3 exploration methodologies, depicting the average and standard deviation over 100 procedurally generated environments. Red and blue colors correspond to the non-learning approaches, while purple color corresponds to the performance of the PPO trained policy. Line type (solid or dashed) denotes the terrain size ($42^2$ or $84^2$).}
		\label{fig:scale}
	\end{figure}

	The analysis is concluded with a scalability study and comparison to non-learning methodologies. Two terrains with sizes $[42\times42]$ and $[84\times84]$ were used. The difficulty level was set to $[d_t, d_m, d_b]=[2,2,1]$, while 100 experiments were conducted for each scenario. Utility and cost-based frontier cell exploration methodologies \cite{basilico2011exploration} were enabled for positioning the achieved PPO policy in the context of non-learning approaches. In these frontier-based approaches, the exploration policy is divided into two categories based on the metric to be optimized:
	
	\textbullet \textit{ Cost:} the next action is chosen based on the distance from the nearest frontier cell.
	
	\textbullet \textit{ Utility:} the decision-making is governed by frequently updated information potential field. 
	
	Figure \ref{fig:scale} summarizes the result of such evaluation study by presenting the average exploration time for each algorithm (PPO, cost frontier-based, utility frontier-based) over 100 procedurally generated runs. A  direct outcome is that the learning-based approach requires the robot to travel less distance to explore the same percentage of terrain as the non-learning approaches. The final remark is devoted to the ``knee'' that can be observed in almost all the final stages of the non-learning approaches. Such behavior is attributed to having several distant sub-parts of the terrain unexplored, the exploration of which requires this extra effort. On the contrary, the learning-based approach (PPO) seems to handle this situation quite well, not leaving these expensive-to-revisit regions along its exploration path.
	
	\section{Conclusions}
	\label{sec:conclusions}
	A new openai-gym environment called MarsExplorer that bridges the gap between reinforcement learning and the real-life exploration/coverage in the robotics domain is presented. The environment transforms the well-known robotics problem of exploration/coverage of a completely unknown region into a reinforcement learning setup that can be tackled by a wide range of off-the-shelf, model-free RL algorithms. An essential feature of the whole solution is that trained policies can be straightforwardly applied to real-life robotic platforms without being trained/tuned to the robot's dynamics. To achieve that, the same level of information abstraction between the robotic system and the MarsExplorer is required. A detailed experimental evaluation was also conducted and presented. 4 state-of-the-art RL algorithms, namely A3C, PPO, Rainbow, and SAC, were evaluated in a challenging version of MarsExplorer, and their training results were also compared with the human-level performance for the task at hand. PPO algorithm achieved the best score, which was also 85.8\% of the human-level performance. Then, the PPO algorithm was utilized to study the effect of the multi-dimensional difficulty vector changes in the overall performance. The visualization of the paths for all these difficulty levels revealed a quite important trait. The PPO learned policy has learned to perform a Hilbert curve with the extra ability to avoid any encountered obstacle. Lastly, a scalability study clearly indicates the ability of RL approaches to be extended in larger terrains, where the achieved performance is validated with non-learning, frontier-based explorations strategies.
	
	\begin{ack}
	This project has received funding from the European Commission under the European Union's Horizon 2020 research and innovation programme under grant agreement no 833464 (CREST). Also, we gratefully acknowledge the support of NVIDIA Corporation with the donation of GPUs used for this research.
	\end{ack}
	\clearpage
	\appendix
	\section{Hyperparameters}
	\label{appendx:Hyperparameters}
	
	\begin{table}[h]
		\caption{PPO Hyperparameters}
		\label{tab:PPO}
		\centering
		\begin{tabular}{ c | c  c}
			\hline
			\textbf{Parameter}  & Value & Comments\\
			\hline
			$\gamma$ & 0.95 & Discount factor of the MDP\\
			$\lambda$ & 5e-5 & Learning rate\\
			Critic & True & Used a critic as a baseline\\
			GAE ${l}$ & 0.95 & GAE (lambda) parameter\\
			KL coeff & 0.2 & Initial coefficient for KL divergence\\
			Clip & 0.3 & PPO clip parameter\\
			\hline
		\end{tabular}
	\end{table}
	
	\begin{table}[h]
		\caption{DQN-Rainbow Hyperparameters}
		\label{tab:DQN}
		\centering
		\begin{tabular}{ c | c  c}
			\hline
			\textbf{Parameter}  & Value & Comments\\
			\hline
			$\gamma$ & 0.95 & Discount factor of the MDP\\
			$\lambda$ & 5e-4 & Learning rate\\
			
			Noisy Net & True & Used a noisy network\\
			Noisy $\sigma$ & 0.5 & initial value of noisy nets\\
			
			Dueling Net & True & Used dueling DQN\\
			Double dueling & True & Used double DQN\\
			
			$\epsilon$-greedy & [1.0, 0.02] & Epsilon greedy for exploration.\\
			
			Buffer size & 50000 & Size of the replay buffer\\
			Priorited Replay & True & Prioritized replay buffer used\\
			
			\hline
		\end{tabular}
	\end{table}
	
	\begin{table}[h]
		\caption{SAC Hyperparameters}
		\label{tab:SAC}
		\centering
		\begin{tabular}{ c | c  c}
			\hline
			\textbf{Parameter}  & Value & Comments\\
			\hline
			$\gamma$ & 0.95 & Discount factor of the MDP\\
			$\lambda$ & 3e-4 & Learning rate\\
			
			Twin Q & True & Use two Q-networks \\
			Q hidden & [256, 256] & Hidden layer activation\\
			
			Policy hidden & [256, 256] & Hidden layer activation\\
			
			Buffer size & 1e6 &  Size of the replay buffer\\
			Priorited Replay & True & Prioritized replay buffer used\\
			
			\hline
		\end{tabular}
	\end{table}
	
	\begin{table}[!h]
		\caption{A3C Hyperparameters}
		\label{tab:A3C}
		\centering
		\begin{tabular}{ c | c  c}
			\hline
			\textbf{Parameter}  & Value & Comments\\
			\hline
			$\gamma$ & 0.95 & Discount factor of the MDP\\
			$\lambda$ & 1e-4 & Learning rate\\
			
			Critic & True & Used a critic as a baseline\\
			GAE & True & General Advantage Estimation\\
			GAE ${l}$ & 0.99 & GAE(lambda) parameter\\
			
			Value loss & 0.5 & Value Function Loss coefficient\\
			Entropy coef & 0.01 & Entropy coefficient\\
			
			
			\hline
		\end{tabular}
	\end{table}
	
	\bibliographystyle{unsrt}
	\bibliography{report}
	
\end{document}